\documentclass[10pt,twocolumn,letterpaper]{article}

\usepackage{cvpr}              
\usepackage{algorithm}
\usepackage{algorithmic}
\usepackage{amsmath}
\usepackage{amsfonts}
\usepackage{amsthm}
\usepackage{listings}
\usepackage{multirow}
\usepackage{booktabs}
\definecolor{cvprblue}{rgb}{0.21,0.49,0.74}
\usepackage[pagebackref,breaklinks,colorlinks,allcolors=cvprblue]{hyperref}

\def\paperID{*****} 
\def\confName{CVPR}
\def\confYear{2025}

\title{Bringing Clustering to MLL: Weakly-Supervised Clustering for Partial Multi-Label Learning}

\author{Yu Chen, Weijun Lv, Yue Huang, Xuhuan Zhu, Fang Li\
\\[0.2em]
School of Automation, Guangdong University of Technology\\
{\tt\small chenyu9265324@163.com}
}

\begin{document}
\maketitle

\begin{abstract}
Label noise in multi-label learning (MLL) poses significant challenges for model training, particularly in partial multi-label learning (PML) where candidate labels contain both relevant and irrelevant labels. While clustering offers a natural approach to exploit data structure for noise identification, traditional clustering methods cannot be directly applied to multi-label scenarios due to a fundamental incompatibility: clustering produces membership values that sum to one per instance, whereas multi-label assignments require binary values that can sum to any number. We propose a novel weakly-supervised clustering approach for PML (WSC-PML) that bridges clustering and multi-label learning through membership matrix decomposition. Our key innovation decomposes the clustering membership matrix $\mathbf{A}$ into two components: $\mathbf{A}=\mathbf{\Pi}\odot\mathbf{F}$, where $\mathbf{\Pi}$ maintains clustering constraints while $\mathbf{F}$ preserves multi-label characteristics. This decomposition enables seamless integration of unsupervised clustering with multi-label supervision for effective label noise handling. WSC-PML employs a three-stage process: initial prototype learning from noisy labels, adaptive confidence-based weak supervision construction, and joint optimization via iterative clustering refinement.  Extensive experiments on 24 datasets demonstrate that our approach outperforms six state-of-the-art methods across all evaluation metrics.
\end{abstract}    
\section{Introduction}\label{sec1}
Multi-label learning \cite{trends} has emerged as a fundamental machine learning paradigm where each instance can be simultaneously associated with multiple class labels. Unlike traditional single-label classification, multi-label learning addresses scenarios where objects naturally belong to multiple categories, such as text categorization where documents may cover multiple topics, image annotation where photos contain multiple objects, or medical diagnosis where patients may have multiple symptoms. This paradigm has been widely applied in multiple fields \cite{mll-app2,mll-app3}.

Traditional multi-label learning approaches assume the availability of complete and accurate label information during training, where each training instance is associated with its exact set of relevant labels. However, this assumption is often unrealistic in practical applications due to the expensive and time-consuming nature of obtaining precise multi-label annotations. Human annotators may miss relevant labels, introduce irrelevant ones, or provide incomplete label sets, making the acquisition of perfect multi-label training data challenging and costly.

\begin{figure}[t]
	\centering
	\includegraphics[width=1\linewidth]{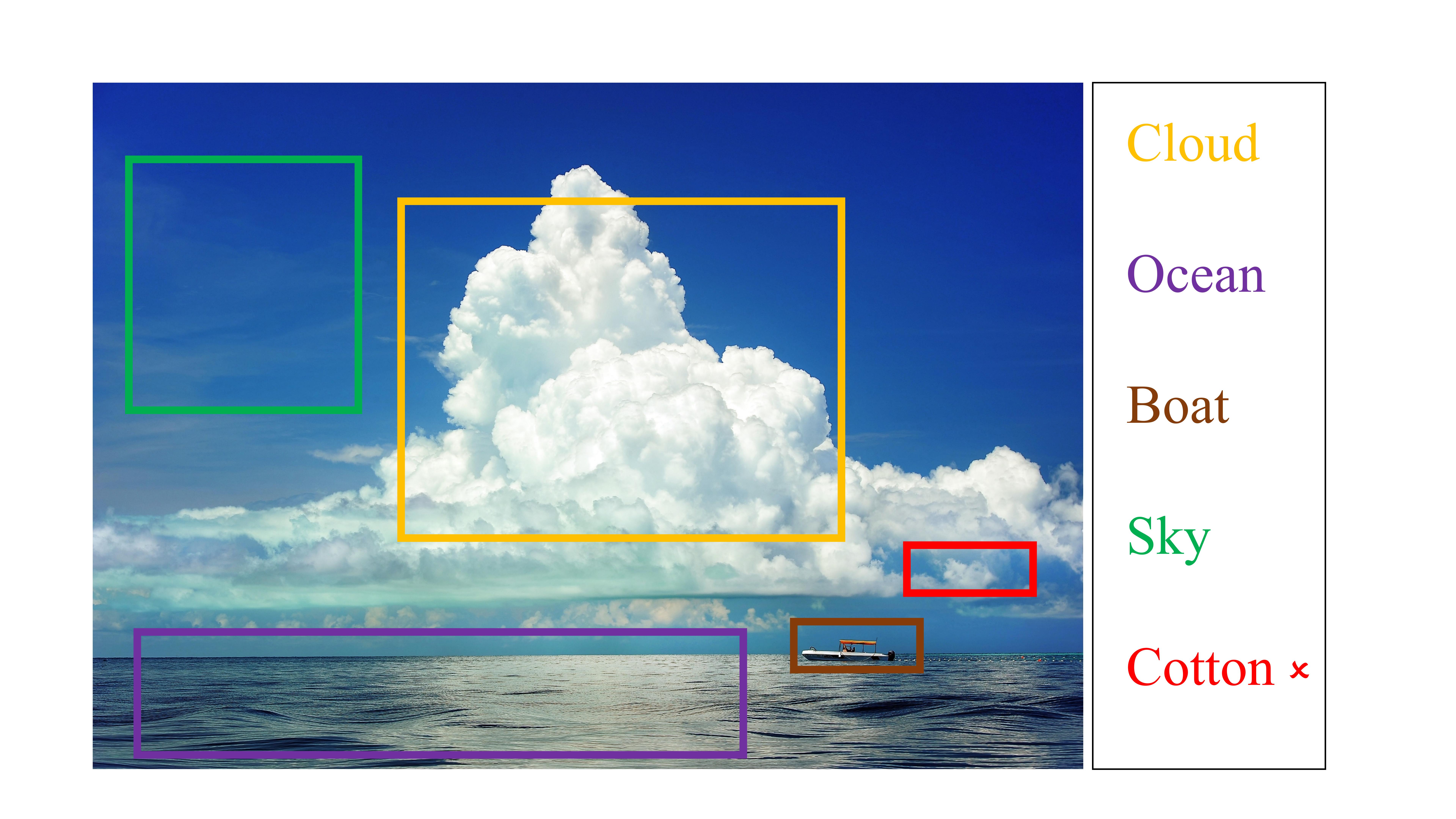}
	\caption{Example of partial multi-label learning with label noise: Candidate labels contain both ground-truth labels (Cloud, Ocean, Boat, Sky) and false positives (Cotton).}
	\label{fig_1}
\end{figure}

To address these limitations, partial multi-label learning (PML) \cite{pml-fp} has been proposed as a more practical learning paradigm. In PML, each training instance is associated with a candidate label set that contains all relevant labels but may also include irrelevant ones. This setting reflects real-world annotation scenarios where it is easier to provide a superset of potential labels rather than the exact label set. For instance, when annotating images, annotators may include all possible objects they suspect to be present, as shown in Figure \ref{fig_1}. While this approach reduces annotation cost and effort, it introduces the significant challenge of learning from noisy label information.

Current approaches to PML predominantly adopt two main paradigms: 1) label correction methods that aim to identify and remove noisy labels from candidate sets \cite{pml-lrs,nlr}; 2) probabilistic modeling approaches that estimate label confidence by exploiting various instance-wise or label-wise relationships \cite{pml-ld,lenfn}. These methods typically focus on developing denoising mechanisms or uncertainty quantification techniques. However, such approaches often overlook the valuable underlying data structure that could provide crucial insights for label disambiguation. Clustering presents a promising avenue for exploiting data structure in PML, as it can naturally group similar instances and potentially separate relevant labels from noise. Some recent work has attempted to integrate clustering with PML \cite{fbd-pml}. However, these approaches face a fundamental limitation: traditional clustering produces membership values that sum to one per instance, but MLL scenarios require binary label assignments, creating an incompatibility that prevents direct use of clustering memberships as multi-label predictions.

To address this incompatibility, we propose WSC-PML, a novel weakly-supervised clustering approach for PML. Our key innovation decomposes the traditional clustering membership matrix $\mathbf{A}$ into two specialized components: $\mathbf{A} = \mathbf{\Pi} \odot \mathbf{F}$, where $\mathbf{\Pi}$ maintains clustering constraints while $\mathbf{F}$ preserves multi-label characteristics . This decomposition enables the first effective integration of unsupervised clustering principles with multi-label supervision, allowing simultaneous exploitation of data structure and label information. Our three-stage approach includes: 1) initial prototype learning from noisy candidate labels, 2) confidence-based weak supervision construction using prototype-distance relationships, and 3) joint optimization of pseudo-labels and class prototypes via weakly-supervised clustering. Through this novel integration, WSC-PML addresses the fundamental limitations of existing PML methods while opening new possibilities for structure-aware multi-label learning.

The main contributions of this paper are as follows:
\begin{itemize}
	\item This paper proposes a novel membership matrix decomposition that resolves the incompatibility between clustering constraints and multi-label scenarios.
	
	\item This paper presents a three-stage weakly-supervised clustering framework that jointly optimizes pseudo-labels and class prototypes.
	
	\item This paper designs an adaptive confidence mechanism that dynamically adjusts supervision strength based on prototype-distance relationships.
\end{itemize}

\begin{figure*}
  \centering
  \includegraphics[width=0.8\linewidth]{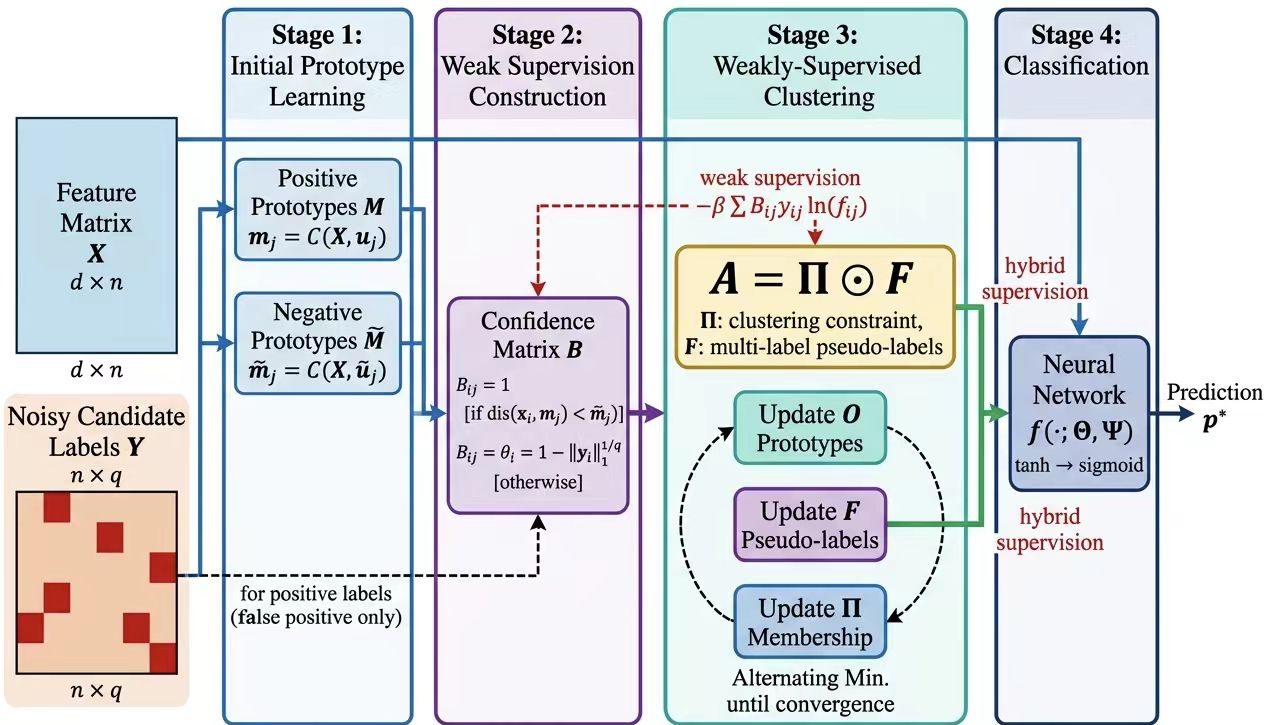}
  \caption{The framework of WSC-PML}
  \label{framework}
\end{figure*}

\section{Related Work}\label{sec2}
Partial multi-label learning is closely related to two popular machine learning frameworks, namely multi-label learning (MLL) and partial label learning (PLL).

Multi-label learning has evolved significantly, addressing scenarios where each instance can be simultaneously associated with multiple class labels. Traditional approaches often converted multi-label problems into binary classification tasks \cite{BR}, but researchers have increasingly focused on exploiting label correlations \cite{CC} and higher-order dependencies \cite{HOMI}. Recent advances have integrated manifold learning principles \cite{mll-1,mll-2}, leveraging the assumption that similar instances may share labels to preserve structures in both feature and label spaces. However, traditional multi-label learning assumes accurate label information, which is unrealistic given the high cost and time requirements of precise annotation.

Partial label learning (PLL) is a multi-class single-label
task where each instance is associated with a set of candidate
labels, one of which is correct and the others are noisy. PLL methods typically employ disambiguation strategies to identify the singular ground-truth label from the candidate set \cite{pll-2}, either by averaging potential label outputs \cite{pll-3} or by treating the ground-truth label as a latent variable in iterative optimization frameworks \cite{pll-1}. PLL algorithms effectively handle noise but are limited to single-label predictions, making them unsuitable for multi-label scenarios where instances belong to multiple classes.

Partial multi-label learning (PML) addresses the challenging scenario where training instances have candidate label sets containing both relevant and irrelevant labels, with false positives being the primary noise source. Current PML research predominantly focuses on label disambiguation—identifying ground-truth labels from noisy candidate sets. Existing disambiguation methods employ diverse techniques: matrix decomposition approaches separate ground-truth from noisy labels using low-rank and sparse decomposition \cite{pml-lrs}; confidence-based methods estimate label credibility through propagation or enhancement \cite{particle,pml-ld,pml-ldl}; feature-driven approaches identify noise by imposing correlation or manifold constraints \cite{fpml,drama,pml-ni,pml-dndc,pml-plr}. Recent advances include feature selection for disambiguation \cite{pml-fsso,lsnrls,lcfs-pml}, label correlation and class prototype exploitation \cite{glc,pml-salc,pml-lc,fbd-pml}, and deep learning for noise reduction \cite{pml-gan,pard,pml-bls}.
\section{Proposed Method}\label{sec3}
In this section, we specifically introduce WSC-PML and its feasibility optimization. Let us first define some common variables as: $\mathbf{X}=[\mathbf{x}_1,\mathbf{x}_2,\cdots,\mathbf{x}_n]\in\mathbb{R}^{d\times n}$ denotes the feature matrix of $n$ instances with $d$-dimensional features. $\mathbf{Y}=[\mathbf{y}_1,\mathbf{y}_2,\cdots,\mathbf{y}_n]^\top\in\{0,1\}^{n\times q}$ represents the candidate label matrix with noisy information, where $q$ is the number of label classes. The training dataset is defined as $\mathcal{D}=\{(\mathbf{x}_i,\mathbf{y}_i)|1\leq i\leq n\}$, containing $n$ labeled instances. If $y_{ij} = 1$, then the $j$-th label is associated with the $i$-th instance; if $y_{ij} = 0$, the association does not exist. Each instance corresponds to a set of candidate labels where unrelated labels are incorrectly labeled as "$1$", which are called noisy labels. The goal of PML is to minimize the impact of noisy information to make correct multi-label predictions.
Our WSC-PML method consists of three main stages:
\begin{enumerate}
	\item \textbf{Initial Prototype Learning with Noisy Labels}: We perform clustering using noisy candidate labels to obtain initial positive and negative class prototypes.
	\item \textbf{Weak Supervision Signal Construction}: We construct reliable weak supervision signals by evaluating label reliability through prototype-based distance comparison.
	\item \textbf{Weakly-Supervised Clustering and classification}: We perform weakly-supervised clustering that jointly optimizes pseudo-labels and class prototypes, followed by final classifier training.
\end{enumerate}
The flowchart of WSC-PML is shown in Figure \ref{framework}.
\subsection{Stage 1: Initial Prototype Learning}\label{subsec:stage1}
In the first stage, we utilize the noisy candidate label matrix $\mathbf{Y}$ to learn initial class prototypes. Despite the presence of noise, the candidate labels still contain valuable information about the underlying label structure. We formulate this as a weighted clustering problem:
\begin{equation}\label{eq:initial_clustering}
	\min_{\mathbf{M}} \sum_{i=1}^{n}\sum_{j=1}^{q}u_{ij}\|\mathbf{x}_{i}-\mathbf{m}_{j}\|_{2}^{2}, ~ \text{where} ~ u_{ij}=\frac{y_{ij}}{\sum_{k=1}^{q}y_{ik}},
\end{equation}
where $\mathbf{M} = [\mathbf{m}_1, \mathbf{m}_2, \cdots, \mathbf{m}_q]$ represents the matrix of class prototypes, and $u_{ij}$ denotes the membership weight of instance $i$ to class $j$. The optimal solution for each class prototype can be derived analytically:
\begin{equation}\label{eq:prototype_calculation}
	\mathbf{m}_j=\mathcal{C}(\mathbf{X},\mathbf{u}_j)=\frac{\sum_{k=1}^n u_{kj}{\mathbf{x}}_k}{\sum_{k=1}^n u_{kj}}, \quad  \forall 1 \leq j \leq q,
\end{equation}
where $\mathcal{C}(\cdot,\cdot)$ denotes the centroid calculation function. This formula computes the weighted average of all instances associated with class $j$ according to the candidate labels.

Simultaneously, we construct negative class prototypes to capture the distributional characteristics of instances that are not associated with each class:
\begin{equation}\label{eq:negative_prototype}
	\mathbf{\tilde{m}}_j=\mathcal{C}(\mathbf{X},\mathbf{\tilde{u}}_j), \text{ where } \tilde{u}_{ij}=\frac{(1-y_{ij})}{\sum_{k=1}^{q}(1-y_{ik})}.
\end{equation}
The negative prototypes $\mathbf{\tilde{m}}_j$ represent centroids of instances not associated with class $j$, enabling distinction between relevant and irrelevant label assignments.
\subsection{Stage 2: Weak Supervision Signal Construction}\label{subsec:stage2}
In the second stage, we construct reliable weak supervision signals to guide the subsequent weakly-supervised clustering.  Due to PML characteristics where false positive labels are the main noise source while negative labels are generally reliable, our construction only focuses on positive labels.

The core principle for construction lies in leveraging the prototype-based distance comparison from Stage 1.  We assume that true positive instances should be closer to positive prototypes than to negative prototypes, while noisy labels (false positives) are more likely to be closer to negative prototypes.  Based on this insight, we construct a confidence matrix $\mathbf{B} \in [0,1]^{n \times q}$ that quantifies the supervision reliability:
\begin{equation}\label{eq:confidence_matrix}
	B_{ij}=\begin{cases}
		1, & \text{if } y_{ij}=1 \text{ and } \mathrm{dis}\left(\mathbf{x}_i,\mathbf{m}_j\right)<\mathrm{dis}\left(\mathbf{x}_i,{\mathbf{\tilde{m}}_j}\right)\\
		\theta_i, & \text{if } y_{ij}=1 \text{ and }\mathrm{dis}\left(\mathbf{x}_i,\mathbf{m}_j\right)\geq\mathrm{dis}\left(\mathbf{x}_i,{\mathbf{\tilde{m}}_j}\right)
	\end{cases},
\end{equation}
where $\mathrm{dis}(\cdot,\cdot)$ denotes the Euclidean distance. Positive instances closer to positive prototypes receive full confidence ($B_{ij} = 1$), while those closer to negative prototypes receive reduced confidence $\theta_i \in (0,1)$. We adopt an adaptive confidence parameter:
\begin{equation}\label{eq:adaptive_confidence}
	\theta_{i} = 1 - \frac{\|\mathbf{y}_{i}\|_{1}}{q}.
\end{equation}
This reflects that instances with more candidate labels are more likely to contain noise, thus receiving weaker supervision signals.   The confidence matrix $\mathbf{B}$ serves as weak supervision signal guiding our clustering framework toward reliable assignments.
\subsection{Stage 3: Weakly-Supervised Clustering}\label{subsubsec:stage3}

In the third stage, we perform weakly-supervised clustering using supervision signals from Stage 2. We begin with a general framework and adapt it for PML characteristics to achieve effective noise correction while maintaining clustering quality.
We first establish a general weakly-supervised clustering objective incorporating weak supervision from $\mathbf{Y}$:
\begin{equation}\label{eq:wsc}
	\min_{\substack{\mathbf{O},\mathbf{A}\geq \mathbf{0},\mathbf{A} \mathbf{1}_{q}=\mathbf{1}_{n}}} \sum_{i=1}^{n}\sum_{j=1}^{q}\{ a_{ij}\|\mathbf{x}_{i}-\mathbf{o}_{j}\|_{2}^{2} - \lambda B_{ij}u_{ij}\ln(a_{ij})\},
\end{equation}
where $\mathbf{A} \in [0,1]^{n \times q}$ represents the  membership matrix, $\mathbf{O}$ denotes class prototypes, and $u_{ij}$ refers to the Stage 1. The first term enforces feature-based clustering, while the second term provides confidence-weighted weak supervision.

However, the constraint $\mathbf{A} \mathbf{1}_{q}=\mathbf{1}_{n}$ forces each row to sum to one, creating numerical conflicts in multi-label scenarios. For instance, true labels [1,1,0] become membership values [0.57, 0.43, 0], significantly reducing label intensities. To address this, we decompose $\mathbf{A}$ into two components:
\begin{equation}\label{eq:trans}
	\mathbf{A} = \boldsymbol{\Pi}\odot\mathbf{F},\quad \text{s.t.}~0\leq f_{ij} \leq y_{ij},~\forall i,j,
\end{equation}
where $\boldsymbol{\Pi} \in [0,1]^{n \times q}$ is the refined membership matrix and $\mathbf{F} \in [0,1]^{n \times q}$ is the pseudo-labels matrix. This preserves label intensities in $\mathbf{F}$ while maintaining clustering constraints via $\boldsymbol{\Pi}$. The first component of Eq.~\eqref{eq:wsc} is changed to:
\begin{equation}\label{eq:wsc_pml}
	\begin{aligned}
		&\min_{\boldsymbol{\Pi},\mathbf{F},\mathbf{O}} \sum_{i=1}^{n}\sum_{j=1}^{q}\pi_{ij} f_{ij}\|\mathbf{x}_{i}-\mathbf{o}_{j}\|_{2}^{2} + \alpha \sum_{i=1}^{n}\sum_{j=1}^{q} \pi_{ij} |f_{ij} - y_{ij}|^2 \\	
		&\text{s.t.} \quad \mathbf{0}_{n\times q}\leq \mathbf{F}\leq \mathbf{Y},~\boldsymbol{\Pi} \geq \mathbf{0}, ~(\mathbf{\Pi}\odot\mathbf{F}^*) \mathbf{1}_{q}=\mathbf{1}_{n}. 
	\end{aligned}
\end{equation}
The first term enforces feature-based clustering quality, while the second provides weak supervision through label consistency.  $\mathbf{F}^*$ represents the replication value of  $\mathbf{F}$. For the second component of Eq.~\eqref{eq:wsc}, we develop it to:
\begin{equation}\label{eq:pseudolabel_learning}
	\min_{\mathbf{F}}\sum_{i=1}^{n}\sum_{j=1}^{q}-B_{ij}y_{ij}\ln(f_{ij}), \text{s.t.}~ 0\leq f_{ij}\leq y_{ij},\forall i,j.
\end{equation}
Finally, we integrate these two components to obtain the final WSC-PML framework:
\begin{equation}\label{eq:main_objective}
	\begin{aligned}
		&\begin{aligned}
			\min_{\boldsymbol{\Pi},\mathbf{F},\mathbf{O}} &\sum_{i=1}^{n}\sum_{j=1}^{q}\pi_{ij} f_{ij}\|\mathbf{x}_{i}-\mathbf{o}_{j}\|_{2}^{2} + \alpha \sum_{i=1}^{n}\sum_{j=1}^{q} \pi_{ij} |f_{ij} - y_{ij}|^2 \\
			&- \beta  \sum_{i=1}^{n}\sum_{j=1}^{q}B_{ij}y_{ij}\ln(f_{ij})	
		\end{aligned}\\
		&\text{s.t.} \quad \mathbf{0}_{n\times q}\leq \mathbf{F}\leq \mathbf{Y},~\boldsymbol{\Pi} \geq \mathbf{0}, ~(\mathbf{\Pi}\odot\mathbf{F}^*) \mathbf{1}_{q}=\mathbf{1}_{n}, 
	\end{aligned}
\end{equation}
where parameters $\alpha$ and $\beta$ control the trade-off between label consistency and confidence-guided supervision.
\subsubsection{Final Classification}\label{subsubsec:classification}
After obtaining the refined pseudo-labels $\mathbf{F}$, we train a neural network classifier using the training dataset $\mathcal{D}=\{(\mathbf{x}_i,\mathbf{y}_i)|1\leq i\leq n\}$ and $\mathbf{F}$:
\begin{equation}\label{eq:neural_network}
	\begin{aligned}
		&\min_{\Theta,\Psi}\mathcal{L}^{\mathrm{train}}(\mathcal{D},\mathbf{F};\Theta,\Psi)\\
		&\quad\quad\quad=-\frac{1}{n}\sum_{z=1}^{n}\sum_{k=1}^{q}\mathcal{H}(\mathbf{f}_z,\mathbf{y}_z,{s}({t}(\mathbf{x}_z,\theta_k),\psi_k)),
	\end{aligned}
\end{equation}
where $t(\mathbf{a};\theta)=\tanh(\mathbf{wa}+\mathbf{b})$ and $s(\mathbf{a};\psi)=\text{sigmoid}(\mathbf{w'a}+\mathbf{b'})$ with parameter sets $\theta=\{\mathbf{w}, \mathbf{b}\}$, $\psi=\{\mathbf{w'}, \mathbf{b'}\}$, $\Theta=\{\theta_k\}_{k=1}^q$ and $\Psi=\{\psi_k\}_{k=1}^q$. The hybrid loss function $\mathcal{H}(f,y,p)=f\ln({p})+(1-y)\ln(1-{p})$ uses refined pseudo-labels $f$ for positive samples and original candidate labels $y$ for negative samples, effectively addressing PML's asymmetric noise characteristics. For test instances $\mathbf{x}^*$, predictions are obtained via $\mathbf{p}^* = s(t(\mathbf{x}^*, \Theta), \Psi)$.
\subsection{Optimization Algorithm}\label{subsec:optimization}
The optimization of Equation~\eqref{eq:main_objective} is non-convex due to the coupling between variables $\boldsymbol{\Pi}$, $\mathbf{F}$, and $\mathbf{O}$. We adopt an alternating minimization approach:\\
\textbf{Updating O (Class Prototypes)}\\
With fixed $\boldsymbol{\Pi}$ and $\mathbf{F}$, the optimal class prototypes can be computed as:

\begin{equation}\label{eq:update_prototypes}
	\mathbf{o}_j = \frac{\sum_{i=1}^{n} \pi_{ij} f_{ij} \mathbf{x}_i}{\sum_{i=1}^{n} \pi_{ij} f_{ij}}, \quad j = 1, 2, \ldots, q.
\end{equation}
\textbf{Updating $\mathbf{F}$ (Pseudo-Labels)}\label{subsubsec:update_f}\\
With fixed $\boldsymbol{\Pi}$ and $\mathbf{O}$, the pseudo-label update involves solving a constrained optimization problem for each element $f_{ij}$. Following the objective function, the sub-problem for each $f_{ij}$ becomes:
\begin{equation}\label{eq:update_pseudolabels}
	\min_{f_{ij}} \pi_{ij} f_{ij}\|\mathbf{x}_{i}-\mathbf{o}_{j}\|_{2}^{2} + \alpha \pi_{ij} (f_{ij} - y_{ij})^2 - \beta B_{ij}y_{ij}\ln(f_{ij}),
\end{equation}
where, subject to $0 \leq f_{ij} \leq y_{ij}$.
Taking the derivative with respect to $f_{ij}$ and setting it to zero, we obtain:
\begin{equation}\label{eq:pseudolabel_derivative}
	\pi_{ij}\|\mathbf{x}_{i}-\mathbf{o}_{j}\|_{2}^{2} + 2\alpha \pi_{ij} (f_{ij} - y_{ij}) - \beta B_{ij}y_{ij} \frac{1}{f_{ij}} = 0.
\end{equation}
Multiplying through by $f_{ij}$ and rearranging:
\begin{equation}
	2\alpha \pi_{ij} f_{ij}^2 + (\pi_{ij}\|\mathbf{x}_{i}-\mathbf{o}_{j}\|_{2}^{2} - 2\alpha \pi_{ij} y_{ij}) f_{ij} - \beta B_{ij}y_{ij} = 0
\end{equation}
This quadratic equation in $f_{ij}$ has the solution:
\begin{equation}\label{eq:pseudolabel_solution}
	\hat{f}_{ij} = \frac{-(a_1) + \sqrt{a_1^2 - 4a_0a_2}}{2a_0},
\end{equation}
where $a_0 = 2\alpha\pi_{ij}$, $a_1 = \pi_{ij}\|\mathbf{x}_{i}-\mathbf{o}_{j}\|_{2}^{2} - 2\alpha\pi_{ij}y_{ij}$, and $a_2 = -\beta B_{ij}y_{ij}$. We take the positive root which is guaranteed to be positive due to the signs of $a_0 > 0$ and $a_2 < 0$.

The final solution is obtained by projecting this value onto the feasible region $[0, y_{ij}]$:
\begin{equation}\label{eq:pseudolabel_final}
	f_{ij} = \max\left(0, \min\left(y_{ij}, \hat{f}_{ij}\right)\right).
\end{equation}
\textbf{Updating $\boldsymbol{\Pi}$ (Membership Matrix)}\label{subsubsec:update_pi}\\ 
When let $\mathbf{F}^*=\mathbf{F}$ and fixing $\mathbf{O}$, the objective reduces to:
\begin{equation}
	\min_{\boldsymbol{\Pi}} \sum_{i=1}^{n}\sum_{j=1}^{q}\pi_{ij} d_{ij}, \quad \text{s.t.} \quad \sum_{j=1}^{q}\pi_{ij}f^*_{ij} = 1, \quad \pi_{ij} \geq 0,
\end{equation}
where $d_{ij} = f_{ij}\|\mathbf{x}_{i}-\mathbf{o}_{j}\|_{2}^{2} + \alpha |f_{ij} - y_{ij}|^2$.

Using Lagrange multipliers with multiplier $\lambda_i$, we get $\frac{\partial \mathcal{L}}{\partial \pi_{ij}} = d_{ij} - \lambda_i f_{ij} = 0$, yielding $\pi_{ij} = \frac{\lambda_i f_{ij}}{d_{ij}}$. From the constraint $\sum_{j=1}^{q}\pi_{ij}f_{ij} = 1$, we obtain:
\begin{equation}\label{eq:opt_Pi}
	\pi_{ij} = \frac{f_{ij}/d_{ij}}{\sum_{k=1}^{q} f_{ik}/d_{ik}}, \quad \forall i,j.
\end{equation}
\begin{algorithm}[t]
	\caption{WSC-PML Algorithm}
	\label{alg:wsc_pml}
	\renewcommand{\algorithmicrequire}{\textbf{Input:}}
	\renewcommand{\algorithmicensure}{\textbf{Output:}}
	\begin{algorithmic}[1]
		\REQUIRE Feature matrix $\mathbf{X} \in \mathbb{R}^{d \times n}$, candidate label matrix $\mathbf{Y} \in \{0,1\}^{n \times q}$, parameters $\alpha$, $\beta$.
		\ENSURE Trained classifier and pseudo-labels $\mathbf{F}$
		
		\STATE \textbf{Stage 1: Initial Prototype Learning}
		\FOR{$j = 1$ to $q$}
		\STATE Compute prototype: $\mathbf{m}_j$ and $ \mathbf{\tilde{m}}_j$ with Eq. ~\eqref{eq:prototype_calculation} and \eqref{eq:negative_prototype}
		\ENDFOR
		
		\STATE \textbf{Stage 2: Weak Supervision Signal Construction}
		\FOR{$i = 1$ to $n$} 		
		\FOR{$j = 1$ to $q$}
		\IF{$y_{ij} = 1 ~and~ \|\mathbf{x}_i - \mathbf{m}_j\|_2 < \|\mathbf{x}_i - \mathbf{\tilde{m}}_j\|_2$}
		\STATE $B_{ij} = 1$
		\ELSE
		\STATE $B_{ij} = \theta_{i}= 1 - \frac{\|\mathbf{y}_{i}\|_{1}}{q}$
		\ENDIF
		\ENDFOR
		\ENDFOR
		
		\STATE \textbf{Stage 3: Weakly-Supervised Clustering}
		\STATE Initialize $\mathbf{F}^{(0)} = \mathbf{Y}$, $\mathbf{O}^{(0)} = \mathbf{M}$, $\boldsymbol{\Pi}^{(0)}=\mathbf{U}$
		\FOR{$t = 1$ to $T_{max}$}
		\STATE Update $\mathbf{O}^{(t)}$ using Equation~\eqref{eq:update_prototypes}
		\STATE Update $\mathbf{F}^{(t)}$ using Equation~\eqref{eq:pseudolabel_final}
		\STATE Update $\boldsymbol{\Pi}^{(t)}$ using Equation~\eqref{eq:opt_Pi}
		\IF{convergence criterion met}
		\STATE \textbf{break}
		\ENDIF
		\ENDFOR
		
		\STATE \textbf{Stage 4: Classification Training}
		\STATE Train neural network using $\mathcal{D}$ and $\mathbf{F}$ with Equation~\eqref{eq:neural_network}
		
		\RETURN Trained classifier parameters $\Theta$ and $\Psi$.
	\end{algorithmic}
\end{algorithm}
\section{Experiments}\label{sec4}
\subsection{Experimental Setup}
\subsubsection{Datasets.}
To evaluate our proposed WSC-PML method, we conduct experiments across 24 datasets: 6 real-world PML datasets and 18 synthetic ones. Table \ref{tab1} summarizes the characteristics of all datasets, including the number of instances ($\# Instances$), dimensionality of feature space ($\# Dims$), number of class labels ($\# Labels$), average number of candidate labels ($Avg.\#CLs$), and average number of ground-truth labels ($Avg.\#GLs$). Specifically, the 6 real-world datasets include \textit{Mirflickr}, \textit{Music\_emotion}, \textit{Music\_style}, \textit{YeastBP}, \textit{YeastCC}, and \textit{YeastMF} \footnote{http://palm.seu.edu.cn/zhangml/}. For the synthetic datasets, we generate 18 PML datasets from 6 multi-label datasets (\textit{emotions}, \textit{birds}, \textit{medical}, \textit{image}, \textit{yeast}, and \textit{corel5k} \footnote{http://mulan.sourceforge.net/datasets.html}) by systematically adding false positive labels. For example, in the \textit{emotions} dataset, ground-truth labels average 1.86 ($Avg.\#GLs$) per instance, while after noise addition, candidate labels average 4 ($Avg.\#CLs$), resulting in approximately 2.14 noise labels per instance. This diverse collection of synthetic datasets with varying noise levels facilitates comprehensive evaluation.
\begin{table}[t]
	\centering
	\caption{The basic information of the experimental dataset. The first six are real-world PML datasets, and the last six are MLL datasets used to generate synthetic PML datasets.}
	\label{tab1}
	\renewcommand{\arraystretch}{1}
	\resizebox{1\linewidth}{!}{	
		\begin{tabular}{lccccc}
			\toprule
			Datsets &\#Instances &\#Dims &\#Labels &Avg.\#CLs &Avg.\#GLs \\
			\hline
			Mirflickr &10433&100&7&3.35&1.77\\ 
			Music$\_$emotion &6833&98&11&5.29&2.42  \\
			Music$\_$style &6839&98&10&6.04&1.44\\ 
			YeastBP &6139&6139&217&5.93&5.54\\ 
			YeastCC &6139&6139&50&1.39&1.35\\ 
			YeastMF &6139&6139&39&1.04&1.01\\ 
			\hline
			emotions &593&72&6&3, 4, 5&1.86\\ 
			birds &645&260&19&3, 4, 5&1.01\\ 
			medical &978&1449&45&5, 7, 9&1.25\\ 
			image &2000&294&5&2, 3, 4&1.23\\ 
			yeast &2417&103&14&7, 9, 11&4.24\\ 
			corel5k &5000&499&374&7, 9, 11&3.52\\
			\bottomrule		
	\end{tabular}}
\end{table}
\subsubsection{Comparison methods.}
For comparative analysis, we implemented six established benchmark methods. A brief description of each comparison algorithm is provided below:
\begin{itemize}
	\item \textbf{FBD-PML} \cite{fbd-pml}: A PML method that assists disambiguation through dual-class prototypes in both feature and label spaces, and employs manifold embedding to maintain structural consistency.
	\item \textbf{PML-LENFN} \cite{lenfn}: A two-stage PML method that utilizes both near and far neighbor information to improve label consistency, while incorporating nonlinear properties to enhance the classifier.
	\item \textbf{NLR} \cite{nlr}: A PML method that constructs positive and negative dual classifiers using  non-candidate labels to assist in noisy label selection.
	\item \textbf{PAMB} \cite{pamb}: A two-stage PML technique that employs ECOC for the generation of binary label sets and applies loss-weighted predictions to handle unseen instances effectively.
	\item \textbf{PML-NI} \cite{pml-ni}: A PML method proposed for truth label prediction and noise label recognition by decomposing the prediction model matrix into independent components.
	\item \textbf{PML-fp} \cite{pml-fp}: A PML method that utilizes feature prototype learning to assign confidence scores to each label and designs corresponding weighted ranking loss functions.
\end{itemize} 

All comparison methods use parameters following the recommended settings from their respective papers. In our WSC-PML method, $\alpha$ and $\beta$ are searched in $\{10^{-2},~10^{-1},~10^{0},~10^{1},~10^{2}\}$. For fair comparison, we apply $10$-fold cross-validation on all datasets and report the mean performance and standard deviation for all methods.
\subsubsection{Evaluation Metrics} 
In our experiment, we evaluate the performance of WSC-PML and other state-of-the-art baselines using five multi-label metrics: \textit{Hamming loss}, \textit{Ranking loss}, \textit{One-error}, \textit{Coverage}, and \textit{Average precision}. For the first four metrics, lower values indicate better performance, while higher values are preferred for \textit{Average Precision}. Detailed definitions on these metrics can be found in \cite{metrics}.
\subsection{Experimental results}
\begin{table*}[htbp]
	\centering
		\caption{The predictive performance of each comparison method on \textit{\textbf{Averge precision}} (mean$\pm$std), where the best  performance (the larger the better) is shown in boldface.}
	\label{tab2}
	\resizebox{1\linewidth}{!}{	
		\begin{tabular}{ccccccccc}
			\toprule
			\textbf{Data Set} & \textbf{Avg\#CLS} & \textbf{WSC-PML} & \textbf{FBD-PML} & \textbf{PML-LENFN} & \textbf{NLR} & \textbf{PAMB} & \textbf{PML-NI} & \textbf{PML-fp} \\
			\hline
			\textbf{Mirflickr} & \textbf{3.35} & \textbf{0.821±0.008} & 0.786±0.009 & 0.798±0.007 & 0.786±0.008 & 0.807±0.051 & 0.786±0.009 & 0.792±0.005 \\
			\textbf{Music\_emotion} & \textbf{5.29} & \textbf{0.649±0.012} & 0.604±0.013 & 0.610±0.012 & 0.610±0.012 & 0.626±0.011 & 0.608±0.012 & 0.435±0.093 \\
			\textbf{Music\_style} & \textbf{6.04} & \textbf{0.746±0.014} & 0.738±0.015 & 0.731±0.016 & 0.739±0.016 & 0.741±0.007 & 0.739±0.015 & 0.472±0.200 \\
			\textbf{YeastBP} & \textbf{5.93} & \textbf{0.436±0.018} & 0.342±0.019 & 0.404±0.015 & 0.401±0.022 & 0.356±0.022 & 0.404±0.022 & 0.286±0.009 \\
			\textbf{YeastCC} & \textbf{1.39} & \textbf{0.608±0.025} & 0.576±0.025 & 0.598±0.013 & 0.569±0.027 & 0.559±0.022 & 0.452±0.026 & 0.360±0.020 \\
			\textbf{YeastMF} & \textbf{1.04} & \textbf{0.490±0.028} & 0.456±0.031 & 0.480±0.012 & 0.470±0.021 & 0.407±0.011 & 0.417±0.013 & 0.359±0.015 \\
			\hline
			\multirow{3}[1]{*}{\textbf{emotions}} & \textbf{3} & 0.805±0.039 & 0.776±0.034 & 0.781±0.033 & 0.780±0.030 & \textbf{0.810±0.017} & 0.777±0.028 & 0.675±0.005 \\
			& \textbf{4} & \textbf{0.791±0.029} & 0.762±0.034 & 0.768±0.028 & 0.740±0.032 & 0.783±0.036 & 0.749±0.034 & 0.581±0.003 \\
			& \textbf{5} & \textbf{0.756±0.034} & 0.698±0.038 & 0.704±0.031 & 0.670±0.040 & 0.750±0.028 & 0.680±0.039 & 0.563±0.002 \\		\hline
			\multirow{3}[1]{*}{\textbf{birds}} & \textbf{3} & \textbf{0.629±0.059} & 0.625±0.061 & 0.627±0.061 & 0.626±0.067 & 0.589±0.052 & 0.617±0.057 & 0.424±0.004 \\
			& \textbf{4} & \textbf{0.597±0.050} & 0.576±0.045 & 0.581±0.048 & 0.575±0.039 & 0.564±0.044 & 0.572±0.041 & 0.418±0.001 \\
			& \textbf{5} & \textbf{0.592±0.054} & 0.568±0.029 & 0.570±0.025 & 0.558±0.031 & 0.495±0.029 & 0.564±0.034 & 0.457±0.023 \\		\hline
			\multirow{3}[0]{*}{\textbf{medical}} & \textbf{5} & \textbf{0.875±0.031} & 0.850±0.025 & 0.848±0.026 & 0.836±0.029 & 0.776±0.020 & 0.835±0.024 & 0.797±0.002 \\
			& \textbf{7} & \textbf{0.860±0.030} & 0.821±0.031 & 0.826±0.034 & 0.785±0.034 & 0.751±0.029 & 0.791±0.036 & 0.846±0.011 \\
			& \textbf{9} & \textbf{0.844±0.031} & 0.786±0.036 & 0.807±0.029 & 0.735±0.043 & 0.721±0.017 & 0.748±0.040 & 0.815±0.016 \\		\hline
			\multirow{3}[0]{*}{\textbf{image}} & \textbf{2} & \textbf{0.814±0.020} & 0.775±0.021 & 0.780±0.024 & 0.772±0.022 & 0.798±0.024 & 0.770±0.020 & 0.672±0.007 \\
			& \textbf{3} & \textbf{0.792±0.019} & 0.743±0.026 & 0.747±0.029 & 0.724±0.025 & 0.748±0.019 & 0.732±0.024 & 0.688±0.009 \\
			& \textbf{4} & \textbf{0.754±0.022} & 0.669±0.015 & 0.671±0.018 & 0.648±0.012 & 0.711±0.026 & 0.653±0.011 & 0.573±0.019 \\		\hline
			\multirow{3}[0]{*}{\textbf{yeast}} & \textbf{7} & \textbf{0.758±0.019} & 0.749±0.019 & 0.753±0.018 & 0.742±0.017 & 0.756±0.014 & 0.746±0.017 & 0.736±0.005 \\
			& \textbf{9} & \textbf{0.750±0.019} & 0.733±0.016 & 0.740±0.017 & 0.719±0.017 & 0.748±0.013 & 0.725±0.016 & 0.727±0.020 \\
			& \textbf{11} & 0.739±0.015 & 0.702±0.013 & 0.721±0.016 & 0.686±0.015 & \textbf{0.741±0.012} & 0.692±0.013 & 0.700±0.003 \\		\hline
			\multirow{3}[1]{*}{\textbf{corel5k}} & \textbf{7} & \textbf{0.306±0.012} & 0.276±0.013 & 0.279±0.013 & 0.275±0.015 & 0.240±0.019 & 0.280±0.014 & 0.284±0.003 \\
			& \textbf{9} & \textbf{0.301±0.012} & 0.269±0.013 & 0.271±0.013 & 0.273±0.015 & 0.229±0.011 & 0.273±0.013 & 0.201±0.001 \\
			& \textbf{11} & \textbf{0.295±0.011} & 0.263±0.012 & 0.264±0.013 & 0.270±0.014 & 0.227±0.018 & 0.265±0.013 & 0.201±0.000 \\
			\bottomrule
	\end{tabular}}%
\end{table*}%
\begin{figure*}[t]
	\centering
	\hspace{1cm}
	\subfloat[hamming~loss]{\includegraphics[width = 0.3\textwidth]{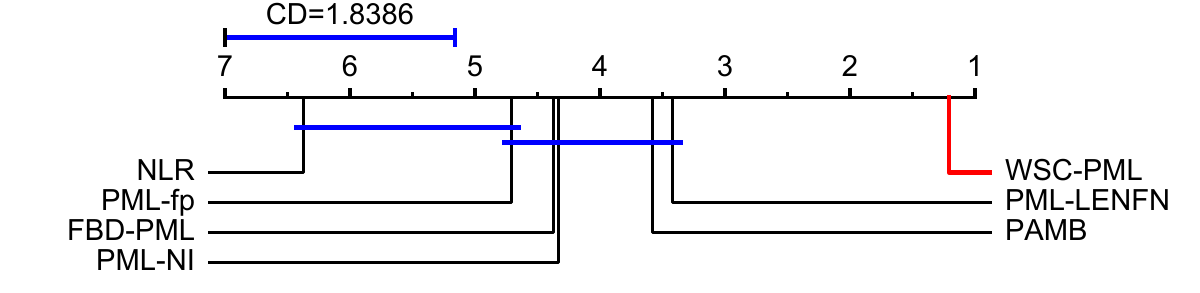}}
	\hspace{2cm}
	\subfloat[ranking~loss]{\includegraphics[width = 0.3\textwidth]{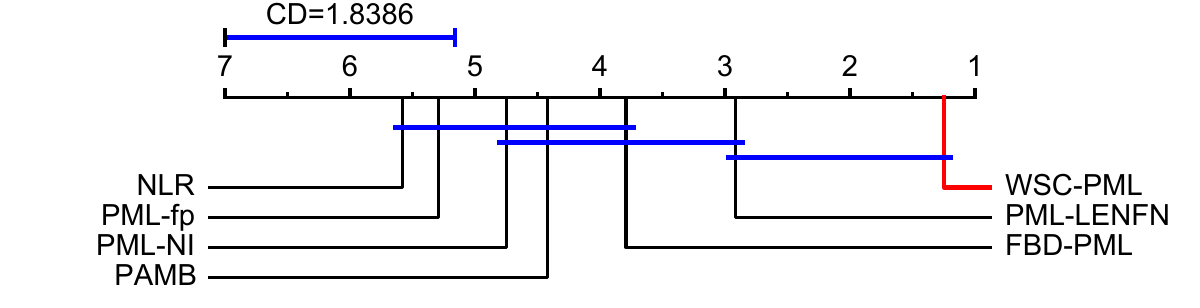}}
	\newline
	\centering
	\vspace{-0.1cm}
	\subfloat[one~error]{\includegraphics[width = 0.3\textwidth]{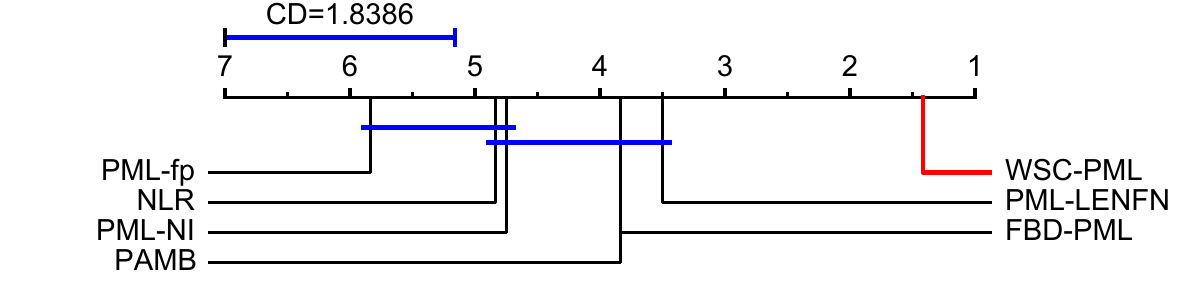}}
	\hfill
	\subfloat[coverage]{\includegraphics[width = 0.3\textwidth]{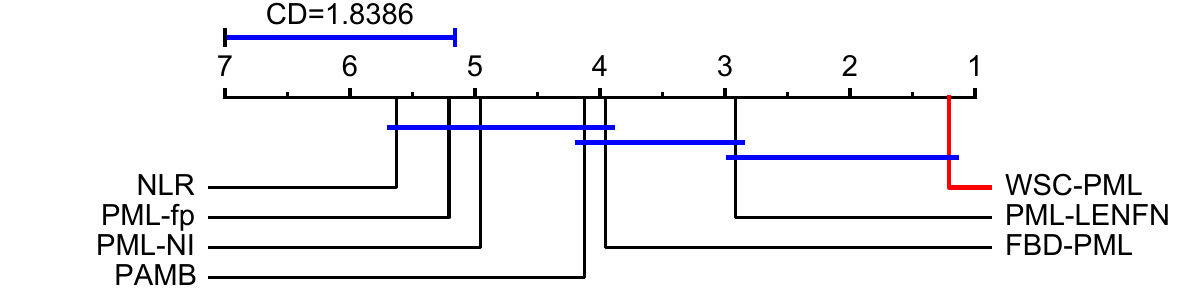}}
	\hfill
	\subfloat[average~precision]{\includegraphics[width = 0.3\textwidth]{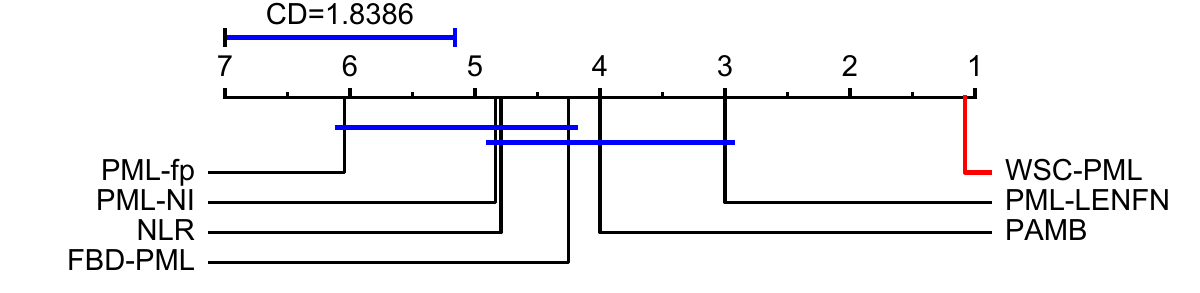}}
	\caption{Results of PML-MA against other approaches with the Nemenyi test(CD = 2.1934 at 0.05 significance level).}
	\label{fig_2}
\end{figure*}
\begin{table}[t]
	\centering
	\caption{Friedman Statistics \( F_F \) for Each Evaluation Metric at 0.05 Significance Level (7 Algorithms and 24 Datasets).}
	\label{tab4}%
	\begin{tabular}{llc}\toprule
		Evaluayion metric & $\ F_F$ & Critical value \\\hline
		Haming loss & 25.4261 &  \\
		Ranking loss & 22.0066 &  \\
		One-error & 16.2616 & 1.8386 \\
		Coverage & 23.0114 &  \\
		Average Precise  & 26.7644 &  \\\bottomrule
	\end{tabular}
\end{table}%
Due to space limitations, we present detailed experimental results for only two evaluation metrics: \textit{Average precision} and \textit{Ranking loss}, as shown in Tables \ref{tab2} and \ref{tab3}. To evaluate statistical significance across all five metrics, we conducted the Friedman test followed by the Nemenyi post-hoc test \cite{demvsar2006statistical}. The Friedman test is a non-parametric test for detecting performance differences across multiple datasets. As shown in Table \ref{tab4}, all Friedman statistics $F_F$ significantly exceed the critical value (1.8386) at 0.05 significance level.

The Nemenyi test is employed to assess statistical significance of performance differences, with WSC-PML designated as the control method. The difference in average rankings is evaluated using the Critical Difference (CD). Performance differences exceeding the CD value are considered statistically significant. Figure \ref{fig_2} presents CD diagrams for all five metrics, with WSC-PML (highlighted in red) serving as the control method. Methods whose performance differences do not exceed the CD threshold are connected by blue lines, indicating no statistically significant difference.
Through comprehensive analysis of all experimental results, we derive the following key observations:

\begin{itemize}
	\item \textbf{Superior Overall Performance:} Across all evaluation metrics and 24 datasets, WSC-PML achieved the best performance in 82.50\% of cases. This stems from our three-stage framework that transforms noisy candidate labels into valuable weak supervision signals, enabling effective exploitation of data structure for simultaneous clustering and denoising.
	
	\item \textbf{Statistical Significance:} Tables \ref{tab2}, \ref{tab3} and Figure \ref{fig_2} demonstrate WSC-PML's significant advantage, consistently maintaining the lowest average ranking across all metrics with few methods connected at the CD threshold. This superior performance results from our confidence-weighted weak supervision integrated with unsupervised clustering principles.
	
	\item \textbf{Advantage over Prototype-based Methods:} Although FBD-PML and PML-fp also utilize prototype learning, WSC-PML significantly outperforms them. The key advantage lies in our membership matrix decomposition $\mathbf{A} = \mathbf{\Pi} \odot \mathbf{F}$ that explicitly addresses the numerical conflict between clustering membership constraints and multi-label scenarios. While existing methods struggle with membership-label intensity conflicts, our decomposition preserves label intensities while maintaining clustering constraints, enabling more effective label disambiguation.
\end{itemize}
\begin{table*}
	\centering
	\caption{The predictive performance of each comparison method on \textit{\textbf{Ranking loss}} (mean$\pm$std), where the best  performance (the smaller the better) is shown in boldface.}
	\label{tab3}
	\resizebox{1\linewidth}{!}{
		\begin{tabular}{ccccccccc}
			\toprule
			\textbf{Data Set} & \textbf{Avg\#CLS} & \textbf{WSC-PML} & \textbf{FBD-PML} & \textbf{PML-LENFN} & \textbf{NLR} & \textbf{PAMB} & \textbf{PML-NI} & \textbf{PML-fp} \\
			\hline
			\textbf{Mirflickr} & \textbf{3.35} & \textbf{0.110±0.006} & 0.124±0.005 & 0.118±0.005 & 0.127±0.006 & 0.112±0.038 & 0.126±0.007 & 0.124±0.003 \\
			\textbf{Music\_emotion} & \textbf{5.29} & \textbf{0.224±0.009} & 0.247±0.009 & 0.246±0.009 & 0.248±0.008 & 0.234±0.007 & 0.246±0.008 & 0.458±0.114 \\
			\textbf{Music\_style} & \textbf{6.04} & \textbf{0.132±0.011} & 0.137±0.011 & 0.138±0.010 & 0.138±0.010 & 0.135±0.005 & 0.137±0.010 & 0.445±0.231 \\
			\textbf{yeastBP} & \textbf{5.93} & \textbf{0.199±0.009} & 0.251±0.009 & 0.226±0.011 & 0.228±0.011 & 0.230±0.011 & 0.220±0.011 & 0.269±0.012 \\
			\textbf{YeastCC} & \textbf{1.39} & \textbf{0.156±0.016} & 0.182±0.020 & 0.178±0.021 & 0.189±0.020 & 0.187±0.026 & 0.193±0.028 & 0.187±0.023 \\
			\textbf{YeastMF} & \textbf{1.04} & \textbf{0.209±0.019} & 0.245±0.022 & 0.235±0.020 & 0.239±0.012 & 0.253±0.025 & 0.247±0.031 & 0.249±0.024 \\
			\hline
			\multirow{3}[2]{*}{\textbf{emotions}} & \textbf{3} & 0.164±0.035 & 0.187±0.032 & 0.180±0.031 & 0.185±0.033 & \textbf{0.160±0.021} & 0.188±0.029 & 0.293±0.015 \\
			& \textbf{4} & \textbf{0.169±0.029} & 0.199±0.029 & 0.192±0.027 & 0.217±0.027 & 0.177±0.032 & 0.211±0.027 & 0.407±0.014 \\
			& \textbf{5} & 0.204±0.029 & 0.260±0.037 & 0.249±0.030 & 0.288±0.040 & \textbf{0.203±0.025} & 0.276±0.039 & 0.419±0.004 \\
			\hline
			\multirow{3}[2]{*}{\textbf{birds}} & \textbf{3} & \textbf{0.172±0.026} & 0.173±0.031 & 0.174±0.033 & 0.183±0.035 & 0.196±0.042 & 0.177±0.033 & 0.271±0.005 \\
			& \textbf{4} & 0.195±0.041 & \textbf{0.194±0.031} & 0.197±0.032 & 0.202±0.029 & 0.204±0.028 & 0.205±0.034 & 0.279±0.024 \\
			& \textbf{5} & \textbf{0.197±0.037} & 0.207±0.031 & 0.204±0.036 & 0.220±0.037 & 0.229±0.025 & 0.219±0.036 & 0.271±0.010 \\
			\hline
			\multirow{3}[2]{*}{\textbf{medical}} & \textbf{5} & \textbf{0.030±0.012} & 0.036±0.013 & 0.036±0.013 & 0.039±0.012 & 0.080±0.021 & 0.040±0.012 & 0.050±0.002 \\
			& \textbf{7} & \textbf{0.033±0.010} & 0.045±0.014 & 0.043±0.014 & 0.052±0.014 & 0.091±0.022 & 0.052±0.013 & 0.040±0.006 \\
			& \textbf{9} & \textbf{0.039±0.013} & 0.052±0.014 & 0.049±0.014 & 0.063±0.015 & 0.104±0.020 & 0.061±0.015 & 0.052±0.004 \\
			\hline
			\multirow{3}[2]{*}{\textbf{image}} & \textbf{2} & \textbf{0.152±0.018} & 0.187±0.019 & 0.188±0.023 & 0.191±0.020 & 0.177±0.022 & 0.194±0.019 & 0.282±0.011 \\
			& \textbf{3} & \textbf{0.173±0.020} & 0.219±0.023 & 0.212±0.026 & 0.235±0.023 & 0.217±0.015 & 0.230±0.024 & 0.265±0.009 \\
			& \textbf{4} & \textbf{0.205±0.018} & 0.287±0.014 & 0.285±0.017 & 0.309±0.012 & 0.255±0.025 & 0.303±0.010 & 0.380±0.017 \\
			\hline
			\multirow{3}[2]{*}{\textbf{yeast}} & \textbf{7} & \textbf{0.173±0.013} & 0.181±0.013 & 0.178±0.012 & 0.186±0.012 & 0.214±0.008 & 0.184±0.012 & 0.185±0.003 \\
			& \textbf{9} & \textbf{0.180±0.01}6 & 0.197±0.016 & 0.190±0.016 & 0.207±0.016 & 0.211±0.007 & 0.202±0.016 & 0.196±0.017 \\
			& \textbf{11} & \textbf{0.192±0.013} & 0.223±0.013 & 0.209±0.014 & 0.235±0.014 & 0.217±0.008 & 0.231±0.014 & 0.219±0.004 \\
			\hline
			\multirow{3}[2]{*}{\textbf{corel5k}} & \textbf{7} & 0.184±0.008 & 0.221±0.010 & 0.217±0.008 & 0.241±0.010 & 0.222±0.009 & 0.215±0.008 & \textbf{0.173±0.005} \\
			& \textbf{9} & 0.192±0.008 & 0.227±0.009 & 0.224±0.008 & 0.246±0.010 & 0.229±0.009 & 0.223±0.008 & \textbf{0.180±0.003} \\
			& \textbf{11} & 0.197±0.007 & 0.232±0.008 & 0.230±0.007 & 0.249±0.010 & 0.235±0.009 & 0.229±0.007 & \textbf{0.181±0.004} \\
			\bottomrule
	\end{tabular}}%
\end{table*}%
\subsection{Further Analysis}
\subsubsection{Parameter Sensitivity Analysis.}
\begin{figure}[t]
	\centering
	\subfloat[Music\_eomtion]{\includegraphics[width = 0.23\textwidth]{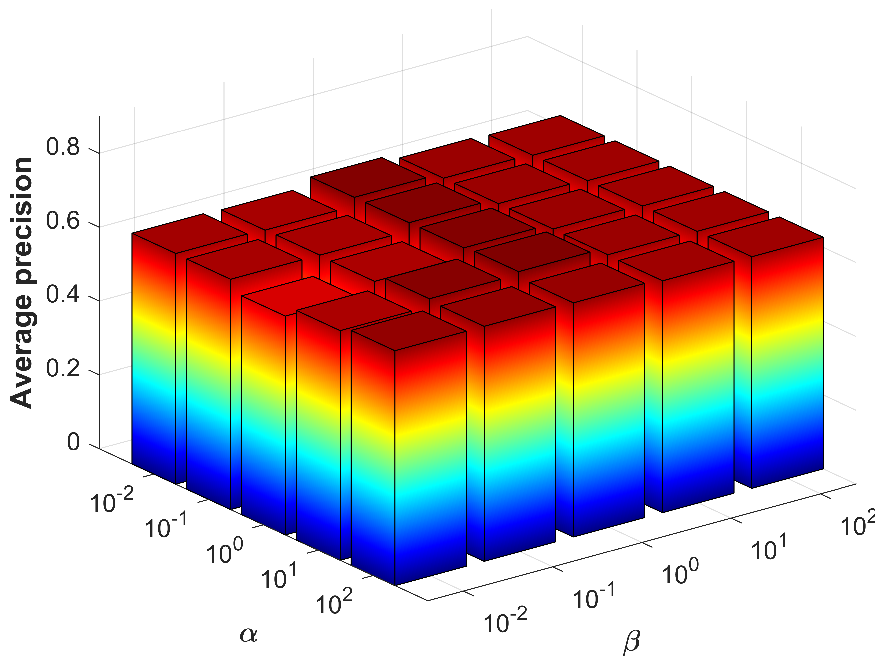}}
	\hfill
	\subfloat[birds (Avg.\#CLs = 4)]{\includegraphics[width = 0.23\textwidth]{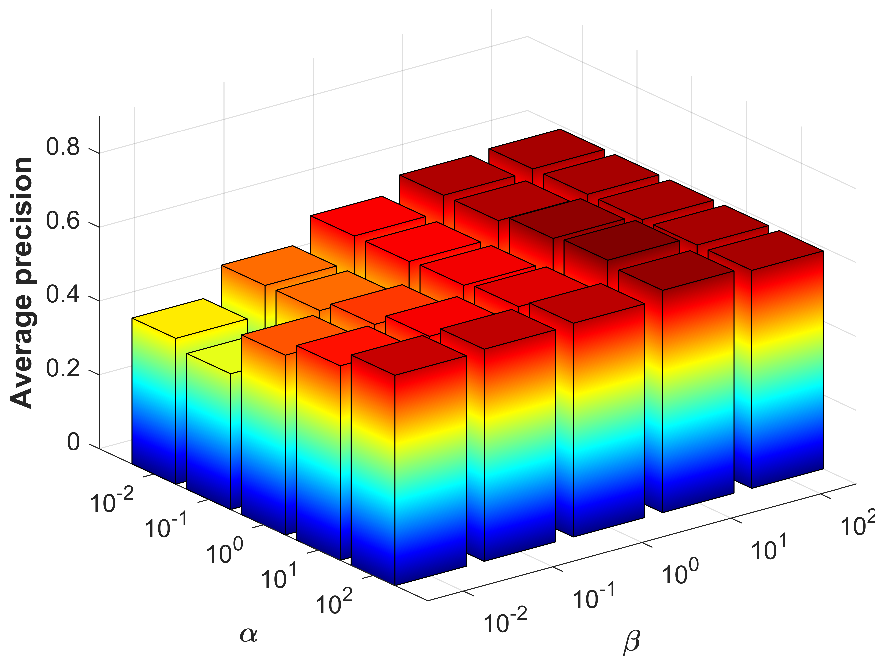}}
	\hfill
	\caption{Parameter Sensitivity Analysis of $\alpha$ and $\beta$ on Mirflickr and birds Datasets}
	\label{fig_3}
\end{figure}

To evaluate the robustness of WSC-PML with respect to hyperparameters $\alpha$ and $\beta$, we conducted comprehensive parameter sensitivity analysis on representative datasets. Figure \ref{fig_3} illustrates the effect of different parameter combinations on Average Precision for Music\_emotion and birds datasets, with both parameters varying in $\{10^{-2}, 10^{-1}, 10^0, 10^1, 10^2\}$. The results demonstrate that WSC-PML achieves optimal performance when both $\alpha$ and $\beta$ are in moderate ranges (approximately $10^{-1}$ to $10^1$), indicating that balanced trade-offs between label consistency and confidence-guided supervision are crucial. Moreover, the extensive high-performance regions (yellow-red areas) across both datasets reveal that our method maintains stable performance over a wide range of parameter combinations, with performance degradation occurring only at extreme values where either supervision signal becomes overly dominant. 
\begin{table*}
	\centering
	\caption{Ablation study results. Comparison of performance with and without each module, in metric \textit{Average precision}. The best experimental performance is shown in boldface.}
	\label{tab5}
	\resizebox{0.8\linewidth}{!}{
		\begin{tabular}{l|cccc}
			\toprule
			{AP $\uparrow$} &NC&ND&NWS&Full \\ \hline
			Mirflickr  & 0.800±0.012 & 0.811±0.012 & 0.815±0.009 &\textbf{0.821±0.008} \\ 
			Music\_emotion  &0.624±0.013&0.627±0.015&0.637±0.019&\textbf{0.649±0.012}\\
			birds (5) &0.584±0.057&0.593±0.044&0.596±0.051&\textbf{0.597±0.050}  \\
			image (3) &0.758±0.018 & 0.781±0.017 & 0.786±0.026&\textbf{0.792±0.019}  \\ \bottomrule
		\end{tabular}
	}
\end{table*}

\subsubsection{Ablation Analysis.}
To validate the effectiveness of key components in WSC-PML, we conduct ablation studies by systematically removing critical modules on four representative datasets (Mirflickr, Music\_emotion, birds, and image):

\begin{itemize}
	\item \textbf{NC} (No Clustering): Removes clustering and only classify using candidate labels.
	\item \textbf{ND} (No Decomposition): Removes membership matrix decomposition $\mathbf{A} = \mathbf{\Pi \odot F}$, using Eq. \eqref{eq:wsc}.
	\item \textbf{NWS} (No Weak Supervision): Removes the weak supervision $- \beta  \sum_{i=1}^{n}\sum_{j=1}^{q}B_{ij}y_{ij}\ln(f_{ij})$
	\item \textbf{Full}: Complete method with all components.
\end{itemize}

As shown in Table \ref{tab5}, the complete WSC-PML achieves the best performance across all datasets, demonstrating synergistic effects among components. NC shows the poorest performance, confirming clustering's essential role in exploiting data structure for label disambiguation. ND exhibits significant degradation, particularly on high-noise datasets, validating that membership matrix decomposition is crucial for handling multi-label numerical conflicts. NWS outperforms NC and ND but falls short of the full method, indicating weak supervision provides valuable refinement. These results confirm that each component contributes meaningfully to the framework, with clustering being fundamental, decomposition being critical for multi-label learning, and weak supervision providing additional performance gains.
\subsubsection{Complexity Analysis.}
The time complexity of WSC-PML is $O(T \cdot nqd)$, where $T$ is the number of iterations. Each iteration involves updating class prototypes $O$ with $O(nqd)$, $F$ with $O(nq)$, and $\Pi$ with $O(nq)$. The space complexity is $O(nd + nq)$ for storing matrices. In practice, WSC-PML converges within 10-20 iterations on most datasets, making it computationally efficient and scalable to large-scale multi-label learning problems.
\section{Conclusion}\label{sec5}
This paper successfully bridges clustering and multi-label learning to address label noise in partial multi-label learning. We resolved the fundamental incompatibility between clustering and multi-label learning by decomposing the membership matrix $\mathbf{A} = \mathbf{\Pi} \odot \mathbf{F}$, where $\mathbf{\Pi}$ maintains clustering constraints while $\mathbf{F}$ preserves multi-label characteristics. Our WSC-PML method enables effective integration of unsupervised clustering with multi-label supervision through a three-stage framework. Extensive experiments demonstrate consistent superiority over state-of-the-art methods. This work transforms clustering and label denoising from separate processes into a unified framework, opening new possibilities for structure-aware multi-label learning.

Future work will explore extending our framework to handle more complex noise patterns and investigating applications to other challenging learning scenarios with noisy supervision.
{
    \small
    \bibliographystyle{ieeenat_fullname}
    \bibliography{main}

\begin{thebibliography}{36}
\providecommand{\natexlab}[1]{#1}
\providecommand{\url}[1]{\texttt{#1}}
\expandafter\ifx\csname urlstyle\endcsname\relax
  \providecommand{\doi}[1]{doi: #1}\else
  \providecommand{\doi}{doi: \begingroup \urlstyle{rm}\Url}\fi

\bibitem[Boutell et~al.(2004)Boutell, Luo, Shen, and Brown]{BR}
Matthew~R Boutell, Jiebo Luo, Xipeng Shen, and Christopher~M Brown.
\newblock Learning multi-label scene classification.
\newblock \emph{Pattern recognition}, 37\penalty0 (9):\penalty0 1757--1771,
  2004.

\bibitem[Chen et~al.(2024)Chen, Wu, Han, Fang, Chen, and Wen]{lenfn}
Yu Chen, Yanan Wu, Na Han, Xiaozhao Fang, Bingzhi Chen, and Jie Wen.
\newblock Partial multi-label learning based on near-far neighborhood label
  enhancement and nonlinear guidance.
\newblock In \emph{Proceedings of the 32nd ACM International Conference on
  Multimedia}, pages 3722--3731, 2024.

\bibitem[Chen et~al.(2025)Chen, Li, Han, Li, Gao, Chan, and Fang]{pml-plr}
Yu Chen, Fang Li, Na Han, Guanbin Li, Hongbo Gao, Sixian Chan, and Xiaozhao
  Fang.
\newblock Pseudo-label reconstruction for partial multi-label learning.
\newblock In \emph{Proceedings of the Thirty-Fourth International Joint
  Conference on Artificial Intelligence, {IJCAI-25}}, pages 4896--4904, 2025.

\bibitem[Dem{\v{s}}ar(2006)]{demvsar2006statistical}
Janez Dem{\v{s}}ar.
\newblock Statistical comparisons of classifiers over multiple data sets.
\newblock \emph{The Journal of Machine learning research}, 7:\penalty0 1--30,
  2006.

\bibitem[Fang et~al.(2025)Fang, Hu, Hu, Chen, Xie, and Han]{fbd-pml}
Xiaozhao Fang, Xi Hu, Yan Hu, Yonghao Chen, Shengli Xie, and Na Han.
\newblock Fuzzy bifocal disambiguation for partial multi-label learning.
\newblock \emph{Neural Networks}, 185:\penalty0 107137, 2025.

\bibitem[Han et~al.(2025)Han, Hu, and Gao]{lcfs-pml}
Qingqi Han, Liang Hu, and Wanfu Gao.
\newblock Integrating label confidence-based feature selection for partial
  multi-label learning.
\newblock \emph{Pattern Recognition}, 161:\penalty0 111281, 2025.

\bibitem[Hang and Zhang(2023)]{pard}
Jun-Yi Hang and Min-Ling Zhang.
\newblock Partial multi-label learning with probabilistic graphical
  disambiguation.
\newblock \emph{Advances in Neural Information Processing Systems},
  36:\penalty0 1339--1351, 2023.

\bibitem[Hao et~al.(2023)Hao, Hu, and Gao]{pml-fsso}
Pingting Hao, Liang Hu, and Wanfu Gao.
\newblock Partial multi-label feature selection via subspace optimization.
\newblock \emph{Information Sciences}, 648:\penalty0 119556, 2023.

\bibitem[Hu et~al.(2023)Hu, Fang, Kang, Chen, Fang, and Xie]{pml-dndc}
Yan Hu, Xiaozhao Fang, Peipei Kang, Yonghao Chen, Yuting Fang, and Shengli Xie.
\newblock Dual noise elimination and dynamic label correlation guided partial
  multi-label learning.
\newblock \emph{IEEE Transactions on Multimedia}, 2023.

\bibitem[Lin et~al.(2025)Lin, Li, Lin, Guo, and Mao]{pml-ldl}
Yaojin Lin, Yulin Li, Shidong Lin, Lei Guo, and Yu Mao.
\newblock Partial multi-label feature selection based on label distribution
  learning.
\newblock \emph{Pattern Recognition}, 164:\penalty0 111523, 2025.

\bibitem[Liu et~al.(2023)Liu, Jia, and Zhang]{pamb}
Bing-Qing Liu, Bin-Bin Jia, and Min-Ling Zhang.
\newblock Towards enabling binary decomposition for partial multi-label
  learning.
\newblock \emph{IEEE transactions on pattern analysis and machine
  intelligence}, 2023.

\bibitem[Liu et~al.(2021)Liu, Wang, Shen, and Tsang]{trends}
Weiwei Liu, Haobo Wang, Xiaobo Shen, and Ivor~W Tsang.
\newblock The emerging trends of multi-label learning.
\newblock \emph{IEEE transactions on pattern analysis and machine
  intelligence}, 44\penalty0 (11):\penalty0 7955--7974, 2021.

\bibitem[Qian et~al.(2024)Qian, Tu, Huang, Shu, and Cheung]{pml-bls}
Wenbin Qian, Yanqiang Tu, Jintao Huang, Wenhao Shu, and Yiu-Ming Cheung.
\newblock Partial multilabel learning using noise-tolerant broad learning
  system with label enhancement and dimensionality reduction.
\newblock \emph{IEEE Transactions on Neural Networks and Learning Systems},
  2024.

\bibitem[Read et~al.(2011)Read, Pfahringer, Holmes, and Frank]{CC}
Jesse Read, Bernhard Pfahringer, Geoff Holmes, and Eibe Frank.
\newblock Classifier chains for multi-label classification.
\newblock \emph{Machine learning}, 85:\penalty0 333--359, 2011.

\bibitem[Si et~al.(2023)Si, Jia, Wang, Zhang, Feng, and Qu]{HOMI}
Chongjie Si, Yuheng Jia, Ran Wang, Min-Ling Zhang, Yanghe Feng, and Chongxiao
  Qu.
\newblock Multi-label classification with high-rank and high-order label
  correlations.
\newblock \emph{IEEE Transactions on Knowledge and Data Engineering},
  36\penalty0 (8):\penalty0 4076--4088, 2023.

\bibitem[Sun et~al.(2019)Sun, Feng, Wang, Lang, and Jin]{pml-lrs}
Lijuan Sun, Songhe Feng, Tao Wang, Congyan Lang, and Yi Jin.
\newblock Partial multi-label learning by low-rank and sparse decomposition.
\newblock In \emph{Proceedings of the AAAI conference on artificial
  intelligence}, pages 5016--5023, 2019.

\bibitem[Sun et~al.(2021)Sun, Feng, Liu, Lyu, and Lang]{glc}
Lijuan Sun, Songhe Feng, Jun Liu, Gengyu Lyu, and Congyan Lang.
\newblock Global-local label correlation for partial multi-label learning.
\newblock \emph{IEEE Transactions on Multimedia}, 24:\penalty0 581--593, 2021.

\bibitem[Tahzeeb and Hasan(2022)]{mll-app2}
Shahab Tahzeeb and Shehzad Hasan.
\newblock A neural network-based multi-label classifier for protein function
  prediction.
\newblock \emph{Engineering, Technology \& Applied Science Research},
  12\penalty0 (1):\penalty0 7974--7981, 2022.

\bibitem[Wang et~al.(2019{\natexlab{a}})Wang, Li, and Zhang]{pll-2}
Deng-Bao Wang, Li Li, and Min-Ling Zhang.
\newblock Adaptive graph guided disambiguation for partial label learning.
\newblock In \emph{Proceedings of the 25th ACM SIGKDD international conference
  on knowledge discovery \& data mining}, pages 83--91, 2019{\natexlab{a}}.

\bibitem[Wang et~al.(2019{\natexlab{b}})Wang, Liu, Zhao, Zhang, Hu, and
  Chen]{drama}
Haobo Wang, Weiwei Liu, Yang Zhao, Chen Zhang, Tianlei Hu, and Gang Chen.
\newblock Discriminative and correlative partial multi-label learning.
\newblock In \emph{IJCAI}, pages 3691--3697, 2019{\natexlab{b}}.

\bibitem[Wang et~al.(2025)Wang, Guan, Xie, Jia, Ye, Duan, and Liang]{pml-lc}
Ke Wang, Yahu Guan, Yunyu Xie, Zhaohong Jia, Hong Ye, Zhangling Duan, and Dong
  Liang.
\newblock Partial multi-label learning with label and classifier correlations.
\newblock \emph{Information Sciences}, 712:\penalty0 122101, 2025.

\bibitem[Xie and Huang(2018)]{pml-fp}
Ming-Kun Xie and Sheng-Jun Huang.
\newblock Partial multi-label learning.
\newblock In \emph{Proceedings of the AAAI conference on artificial
  intelligence}, pages 4302--4309, 2018.

\bibitem[Xie and Huang(2021)]{pml-ni}
Ming-Kun Xie and Sheng-Jun Huang.
\newblock Partial multi-label learning with noisy label identification.
\newblock \emph{IEEE Transactions on Pattern Analysis and Machine
  Intelligence}, 44\penalty0 (7):\penalty0 3676--3687, 2021.

\bibitem[Xu et~al.(2020)Xu, Liu, and Geng]{pml-ld}
Ning Xu, Yun-Peng Liu, and Xin Geng.
\newblock Partial multi-label learning with label distribution.
\newblock In \emph{Proceedings of the AAAI conference on artificial
  intelligence}, pages 6510--6517, 2020.

\bibitem[Xu et~al.(2024)Xu, Qiao, Zhao, Geng, and Zhang]{pll-3}
Ning Xu, Congyu Qiao, Yuchen Zhao, Xin Geng, and Min-Ling Zhang.
\newblock Variational label enhancement for instance-dependent partial label
  learning.
\newblock \emph{IEEE Transactions on Pattern Analysis and Machine
  Intelligence}, 2024.

\bibitem[Yan and Guo(2021)]{pml-gan}
Yan Yan and Yuhong Guo.
\newblock Adversarial partial multi-label learning with label disambiguation.
\newblock In \emph{Proceedings of the AAAI Conference on Artificial
  Intelligence}, pages 10568--10576, 2021.

\bibitem[Yang et~al.(2024)Yang, Jia, Liu, Dong, and Hou]{nlr}
Fuchao Yang, Yuheng Jia, Hui Liu, Yongqiang Dong, and Junhui Hou.
\newblock Noisy label removal for partial multi-label learning.
\newblock In \emph{Proceedings of the 30th ACM SIGKDD Conference on Knowledge
  Discovery and Data Mining}, pages 3724--3735, 2024.

\bibitem[Yu et~al.(2018)Yu, Chen, Domeniconi, Wang, Li, Zhang, and Wu]{fpml}
Guoxian Yu, Xia Chen, Carlotta Domeniconi, Jun Wang, Zhao Li, Zili Zhang, and
  Xindong Wu.
\newblock Feature-induced partial multi-label learning.
\newblock In \emph{2018 IEEE international conference on data mining (ICDM)},
  pages 1398--1403. IEEE, 2018.

\bibitem[Yu et~al.(2024)Yu, Wang, and Zhang]{pll-1}
Xiang-Ru Yu, Deng-Bao Wang, and Min-Ling Zhang.
\newblock Dimensionality reduction for partial label learning: A unified and
  adaptive approach.
\newblock \emph{IEEE Transactions on Knowledge and Data Engineering}, 2024.

\bibitem[Zhang et~al.(2019)Zhang, Luo, Li, Zhou, and Li]{mll-1}
Jia Zhang, Zhiming Luo, Candong Li, Changen Zhou, and Shaozi Li.
\newblock Manifold regularized discriminative feature selection for multi-label
  learning.
\newblock \emph{Pattern Recognition}, 95:\penalty0 136--150, 2019.

\bibitem[Zhang et~al.(2024)Zhang, Li, Shen, He, Tan, and He]{mll-app3}
Jun Zhang, Yubin Li, Fanfan Shen, Yueshun He, Hai Tan, and Yanxiang He.
\newblock Hierarchical text classification with multi-label contrastive
  learning and knn.
\newblock \emph{Neurocomputing}, 577:\penalty0 127323, 2024.

\bibitem[Zhang and Fang(2020)]{particle}
Min-Ling Zhang and Jun-Peng Fang.
\newblock Partial multi-label learning via credible label elicitation.
\newblock \emph{IEEE Transactions on Pattern Analysis and Machine
  Intelligence}, 43\penalty0 (10):\penalty0 3587--3599, 2020.

\bibitem[Zhang and Zhou(2013)]{metrics}
Min-Ling Zhang and Zhi-Hua Zhou.
\newblock A review on multi-label learning algorithms.
\newblock \emph{IEEE transactions on knowledge and data engineering},
  26\penalty0 (8):\penalty0 1819--1837, 2013.

\bibitem[Zhao et~al.(2022{\natexlab{a}})Zhao, Zhao, Zhao, Liu, and
  Ji]{pml-salc}
Peng Zhao, Shiyi Zhao, Xuyang Zhao, Huiting Liu, and Xia Ji.
\newblock Partial multi-label learning based on sparse asymmetric label
  correlations.
\newblock \emph{Knowledge-Based Systems}, 245:\penalty0 108601,
  2022{\natexlab{a}}.

\bibitem[Zhao et~al.(2022{\natexlab{b}})Zhao, Zhang, and Pedrycz]{mll-2}
Tianna Zhao, Yuanjian Zhang, and Witold Pedrycz.
\newblock Robust multi-label classification with enhanced global and local
  label correlation.
\newblock \emph{Mathematics}, 10\penalty0 (11):\penalty0 1871,
  2022{\natexlab{b}}.

\bibitem[Zou et~al.(2024)Zou, Hu, Li, and Ge]{lsnrls}
Yizhang Zou, Xuegang Hu, Peipei Li, and Yuhang Ge.
\newblock Learning shared and non-redundant label-specific features for partial
  multi-label classification.
\newblock \emph{Information Sciences}, 656:\penalty0 119917, 2024.

\end{thebibliography}
}


\end{document}